\pdfoutput=1
\documentclass[11pt]{article}

\usepackage{emnlp2021}

\usepackage{times}
\usepackage{latexsym}
\usepackage{verbatim}

\usepackage[T1]{fontenc}
\usepackage[utf8]{inputenc}

\usepackage{microtype}

\title{Improving Large-scale Language Models and Resources for Filipino}

\author {Jan Christian Blaise Cruz \textnormal{and} Charibeth Cheng \\
  Center for Language Technologies \\
  College of Computer Studies \\
  De La Salle University, Manila \\
  {\tt \{jan\_christian\_cruz, charibeth.cheng\}@dlsu.edu.ph} \\}

\begin{document}
\maketitle

\begin{abstract}
In this paper, we improve on existing language resources for the low-resource Filipino language in two ways. First, we outline the construction of the TLUnified dataset, a large-scale pretraining corpus that serves as an improvement over smaller existing pretraining datasets for the language in terms of scale and topic variety. Second, we pretrain new Transformer language models following the RoBERTa pretraining technique to supplant existing models trained with small corpora. Our new RoBERTa models show significant improvements over existing Filipino models in three benchmark datasets with an average gain of 4.47\% test accuracy across the three classification tasks of varying difficulty.
\end{abstract}

\section{Introduction}
Unlike High-resource Languages (HRL) such as English, German, and French, Low-resource Languages (LRL) suffer from a lack of benchmark datasets, databases, linguistic tools, and pretrained models that impede the progress of research within those languages. 

Despite the growing success of methods that intrinsically learn from little data \cite{deng2020low,lee2021meta}, creating ``more data'' remains a very significant fundamental task in NLP. Given the data-hungry nature of the neural networks that are prevalent in NLP today, creating new datasets to train from is the most efficient way to improve model performance. In addition, cleverly-constructed datasets also reveal new insights into the models we commonly use, letting us gauge their true performance and expose hidden weaknesses \cite{maudslay2021syntactic}.

In this paper, we improve upon the existing resources for Filipino, a low-resource language spoken in the Philippines. We create a larger, more topically-varied large-scale pretraining dataset that improves upon the existing WikiText-TL-39 \cite{cruz2019evaluating} that is too small and too topically-narrow to create robust models that perform well in modern NLP. We also produce new RoBERTa pretrained models using our pretraining dataset that supplant existing models trained with less data \cite{cruz2020establishing}.

\section{Resource Creation}
In this section, we outline our full methodology for resource creation. First, we introduce the construction of our new large-scale pretraining dataset. Next, we detail the pretraining steps for our new RoBERTa models. Lastly, we introduce the task datasets that we use to benchmark performance for our new pretrained models.

\subsection{The TLUnified Dataset}
To effectively pretrain a large transformer for downstream tasks, we require an equally large pretraining corpus of high-quality Filipino text. We construct our pretraining corpus by combining a number of available Filipino corpora, including:

\begin{itemize}
    \item \textbf{Bilingual Text Data} -- Bitext datasets are used for training Machine Translation models and contain crawled and aligned data from multiple sources. We collected multiple bitexts, extracted Filipino text, then deduplicated the extracted data to add to our pretraining corpus. Datasets we collected from include bible-uedin \cite{christodouloupoulos2015massively}, CCAligned \cite{elkishky_ccaligned_2020}, ELRC 2922\footnote{https://elrc-share.eu/}, MultiCCAligned \cite{elkishky_ccaligned_2020},ParaCrawl \footnote{https://www.paracrawl.eu/}, TED2020 \cite{reimers-2020-multilingual-sentence-bert}, WikiMatrix \cite{schwenk2019wikimatrix},  tico-19, Ubuntu, OpenSubtitles, QED, Tanzil, Tatoeba, GlobalVoices, KDE4, and WikiMedia \cite{TIEDEMANN12.463}.
    
    \item \textbf{OSCAR} -- The Open Super-Large Crawled Aggregated Corpus (OSCAR) \cite{OrtizSuarezSagotRomary2019} is a massive dataset obtained from language identification and filtering of the Common Crawl dataset. We use the deduplicated version of the Filipino (Tagalog) portion of OSCAR and add it to our pretraining corpus.
    
    \item \textbf{NewsPH} -- The NewsPH \cite{cruz2021exploiting} corpus is a large-scale crawled corpus of Filipino news articles, originally used in automatically creating the NewsPH-NLI benchmark dataset. Since we plan on using an NLI dataset derived from NewsPH for benchmarking in this paper, we opted to only use a 60\% subset of the NewsPH corpus to add to TLUnified.
\end{itemize}
    
Since a large portion of our corpus is crawled and artificially aligned, we expect that out-of-the-box data quality would be low. To clean our dataset, we apply a number of preprocessing filters to it, including:

\begin{enumerate}
    \item Non-latin Filter -- We filter out sentences whose characters are composed of more than 15\% non-latin letters.
    \item Length Filter -- We remove sentences that have a number of tokens $N$ where $4 <= N <= 150$.
    \item Puncutation Filter -- All sentences that have tokens composed of too many succeeding punctuations (eg. ``///'') are all removed.
    \item Average Word Length Filter -- If a sentence has tokens that are significantly longer than the other tokens in the sentence, we remove the sentence entirely. We first take the sum of the character lengths of each token, then divide it by the number of tokens to get a ratio $r$. Only sentences with ratio $3 <= r <= 18$ are kept in the corpus.
    \item HTML Filter -- All sentences with HTML and URL-related tokens (e.g. ``.com'' or ``http://'') are removed.
\end{enumerate}

After filtering the dataset, we perform one additional deduplication step to ensure that no identical lines are found in the dataset. The final result is a large-scale pretraining dataset we call \textbf{TLUnified}.

We then train tokenizers using TLUnified, limiting our vocabulary to a fixed 32,000 BPE subwords \cite{sennrich2015neural}. Our tokenizers are trained with a character coverage of 1.0. We also do not remove casing to ensure that capitalization is kept after tokenization. 

\subsection{Pretraining}
We then pretrain transformer language models that can serve as bases for a variety of downstream tasks later on. For this purpose, we use the RoBERTa \cite{liu2019roberta} pretraining technique. Previous pretrained transformers in Filipino \cite{cruz2020establishing,cruz2021exploiting} used BERT \cite{devlin2018bert}, and ELECTRA \cite{clark2020electra} as their method of choice. 

We choose RoBERTa as it retains state-of-the-art performance on multiple NLP tasks while keeping its pretraining task simple unlike methods such as ELECTRA. As a reproduction study of BERT, RoBERTa optimizes and builds up on the BERT pretraining scheme to improve training efficiency and downstream performance. 

Two size variants are trained in this study following the original RoBERTa paper: a Base model (110M parameters) and a Large model (330M parameters). Both size variants use the same BPE tokenizer trained with TLUnified. Our hyperparameter choices also follow the original RoBERTa paper closely. A summary of our models' hyperparameters can be found in Table \ref{tab:model_hyperparams}.

\begin{table}[]
    \centering
    \begin{tabular}{lcc}
    \hline
     & Base & Large \\
     \hline
     Hidden Size & 768 & 1024 \\
     Feedforward Size & 3072 & 4096 \\
     Max Sequence Length & 512 & 512 \\
     Attention Heads & 12 & 16 \\
     Hidden Layers & 12 & 24 \\
     Droput & 0.1 & 0.1 \\
     \hline
    \end{tabular}
    \caption{Base and Large RoBERTa hyperparameters.}
    \label{tab:model_hyperparams}
\end{table}

During training, we construct batches by continually filling them with tokens until we reach a maximum batch size of 8192 tokens. Both variants are trained using the Adafactor \cite{shazeer2018adafactor} optimizer with $\beta_2 = 0.98$ and a weight decay of 0.01. The base model is trained for 100,000 steps with a learning rate of 6e-4, while the large variant is trained for 300,000 steps with a learning rate of 4e-4. We also use a learning rate schedule that linearly warms up for 25,000 steps, then linearly decays for the rest of training. All experiments are done on a server with 8x NVIDIA Tesla P100 GPUs.

\subsection{Benchmark Datasets}
We test the efficacy of our RoBERTa models on three Filipino benchmark datasets:

\begin{itemize}
    \item \textbf{Filipino Hatespeech Dataset} -- 10,000 tweets labelled as ``hate'' and ``non-hate'' collected during the 2016 Philippine Presidential Elections. Originally published in \citet{cabasag2019hate} and benchmarked with modern Transformers in \citet{cruz2020establishing}.
    \item \textbf{Filipino Dengue Dataset} -- Low-resource multiclass classification dataset with ~4000 samples that can be one or many of five labels. Originally published in \citet{livelo2018intelligent} and benchmarked in \citet{cruz2020establishing} using pretrained Transformers.
    \item \textbf{NewsPH-NLI} -- An automatically-generated dataset constructed by exploiting the ``inverted-pyramid'' structure of news articles, causing every sentence to naturally entail the sentence that came before it. Originally created in \citet{cruz2021exploiting}. 
\end{itemize}

\begin{table*}[h]
    \centering
    \begin{tabular}{l|cc|cc|cc}
         \hline
         & \multicolumn{2}{c|}{\textbf{Hatespeech}} & \multicolumn{2}{c|}{\textbf{Dengue}} & \multicolumn{2}{c}{\textbf{NewsPH-NLI Med.}} \\
         \multicolumn{1}{c|}{Model} & Val. Acc & Test Acc. & Val. Acc & Test Acc. & Val. Acc & Test Acc. \\
         \hline
         BERT Base Cased & 0.7479 & 0.7417 & 0.7720 & 0.7580 & 0.8838 & 0.8874 \\
         ELECTRA Base Cased & 0.7491 & 0.7250 & 0.7400 & 0.6920 & 0.9094 & 0.9106 \\
         RoBERTa Base & 0.7866 & 0.7807 & 0.8180 & 0.8020 & 0.9492 & 0.9501 \\
         RoBERTa Large & 0.7897 & 0.7824 & 0.8281 & 0.8110 & 0.9499 & 0.9510 \\
         \hline
    \end{tabular}
    \caption{Finetuning results for all Transformer variants on the three benchmark datasets.}
    \label{tab:finetuning_results}
\end{table*}

For this study, we do not use the original NewsPH-NLI created in \citet{cruz2021exploiting} as it has significant overlap with the subset of the NewsPH corpus that we used for pretraining. We instead re-generated a version of NewsPH-NLI (which we call ``NewsPH-NLI Medium'') using 40\% of the NewsPH corpus, using the other 60\% as part of the TLUnified pretraining data. This ensures that no test data is present in the training data, which will significantly inflate the benchmark scores.

Preprocessing for the downstream benchmark datasets is kept simple and non-destructive to preserve the linguistic structures and information present in the original data.

For the Hatespeech and the Dengue datasets, we follow the preprocessing used in \citet{cruz2020establishing}, with a number of changes. Since both are datasets composed mainly of tweet data, the following preprocessing steps are done:

\begin{itemize}
    \item Moses detokenization \cite{koehn2010moses} was applied on all Moses-tokenized text.
    \item All HTML meta text and link texts are collapsed into a special \texttt{[LINK]} token. This is to reduce the noise in the dataset as images in the tweets are naturally converted into links.
    \item All substrings that start with an \texttt{@} character that are greater than length 1 are automatically treated as a ``mention'' and are replaced with a \texttt{[MENTION]} special token.
    \item All substrings that start with a \texttt{\#} character that are greater than length 1 are automatically treated as a ``hashtag'' and are replaced with a \texttt{[HASHTAG]} special token.
    \item We renormalize apostrophes (e.g. \texttt{it 's} $\rightarrow$ \texttt{it's}) and punctuation that were spaced out (e.g. \texttt{one - two} $\rightarrow$ \texttt{one-two}) during the preprocessing in the \citet{cruz2020establishing} paper.
    \item Characters that were converted into unicode (e.g. \texttt{\&amp;}) are converted back into their encoded form (e.g. \texttt{\&}).
\end{itemize}

For the Dengue dataset, we transform the multilabel, multiclass classification setup into a multiclass classification problem by concatenating an example's labels and converting the resulting binary number into an integer. For example, a sentence with the labels \texttt{1, 1, 0, 1, 1} for \texttt{absent}, \texttt{dengue}, \texttt{healthclasses}, \texttt{mosquito}, and \texttt{sick} will be converted into \texttt{27} (\texttt{11011} $\rightarrow$ \texttt{27}). This results in 32 possible labels and increases the difficulty of the task.

For the NewsPH-NLI Medium dataset, we opted to not do any further preprocessing as the released data from \citet{cruz2021exploiting} is already preprocessed and clean.

\begin{comment}
Sample preprocessed data from the Hatespeech, Dengue, and NewsPH-NLI Medium datasets can be found in Figure \ref{fig-sample-preproc}.
\end{comment}

\subsection{Finetuning Setups}
We then finetune for the downstream benchmark tasks using our pretrained RoBERTa models. Since the NewsPH-NLI version and the setup of the Dengue dataset task is different from the previous benchmarking paper, we also finetuned Tagalog BERT \cite{cruz2019evaluating} and Tagalog ELECTRA \cite{cruz2021exploiting} to serve as baseline models against the new RoBERTa model.

\begin{table}[]
    \centering
    \begin{tabular}{lcc}
    \hline
     & Base & Large \\
     \hline
     Max. Seq. Length & 128 & 256 \\
     Learning Rate & 2e-5 & 1e-5 \\
     Warmup Ratio & 0.1 & 0.06 \\
     \hline
    \end{tabular}
    \caption{Unique finetuning hyperparameters for Base and Large transformer variants.}
    \label{tab:finetuning_hyperparams}
\end{table}

All models are trained using the Adafactor \cite{shazeer2018adafactor} optimizer with a learning rate scheduler that linearly increases from zero after a ratio of steps-to-total-training-steps has reached, then linearly decays afterwards. We use a batch size of 32 sentences for all models and use a weight decay of 0.1. We opted to use a larger maximum sequence length for the Large RoBERTa models as it has more capacity due to its deeper encoder stack. Hyperparameters that are different between Base and Large variants of the pretrained Transformers used are found in Table \ref{tab:finetuning_hyperparams}.

We add the \texttt{[LINK]}, \texttt{[MENTION]}, and \texttt{[HASHTAG]} special tokens during finetuning for the Hatespeech and Dengue datasets to the vocabularies of the Transformers used, averaging the vectors of all subword embeddings in the embedding layer to serve as initialization for the three added tokens.

Despite our RoBERTa having a full maximum sequence length allowance of 512, we opted to use smaller maximum sequence lengths during finetuning. This speeds up training (approximately 4x for the Base models and 2x for the Large models) while losing zero information since no sentence or sentence pair in any task reaches 256 subwords in length.

All experiments are done on a server with 8x NVIDIA Tesla P100 GPUs.

\section{Results}
We report the results for our finetuning for the three benchmark datasets in terms of validation and test accuracy. A summary of the results can be found on Table \ref{tab:finetuning_results}.

Our RoBERTa models outperformed both the BERT and the ELECTRA models across all tasks. For the Hatespeech task, RoBERTa Large outperformed the best previous model (BERT Base) by +4.07\% test accuracy. RoBERTa large also had a gain in performance in the Dengue dataset (+5.3\% test accuracy over BERT Base) and the NewsPH-NLI Medium dataset (+4.04\% test accuracy over ELECTRA Base).

While marginally inferior to the Large variant, the Base RoBERTa variant still outperforms the baseline models in all tasks. RoBERTa Base has an improvement of +3.9\% against BERT Base on the Hatespeech task, +4.4\% against BERT Base on the Dengue task, and +3.95\% against ELECTRA Base on the NewsPH-NLI Medium task.

The difference in performance between the Base and Large RoBERTa variants is marginal in the current benchmarks. Large outperforms Base only by +0.17\% for Hatespeech, +0.9\% for Dengue, and +0.09\% for NewsPH-NLI Medium. We hypothesize that this is due to the size of the pretraining dataset. While the size of TLUnified is much larger than the previous WikiText-TL-39, it may still not be enough to make full use of the capacity of a Large-variant Transformer. We surmise that RoBERTa Large may need to be trained with more data to show significant, non-marginal improvements in performance.

Overall, our new models show significant improvements over older pretrained Filipino Transformer models. This is likely due to the improved pretraining corpus, with TLUnified being larger and of more varied topics and sources than the previous WikiText-TL-39.

\section{Conclusion}
Our work has two main contributions in terms of language resources for the Filipino language. First, we construct TLUnified, a new large-scale pretraining corpus for Filipino. This is an improvement over the much smaller pretraining corpora currently available, boasting much larger scale and topic variety. Second, we release new pretrained Transformers using the RoBERTa pretraining method. Our new models outperform existing baselines on three different classification tasks, with significant improvements of +4.07\%, +5.03\%, and +4.04\% test accuracy for the Hatespeech, Dengue, and NewsPH-NLI Medium datasets respectively.

\bibliography{anthology,custom}
\bibliographystyle{acl_natbib}

\end{document}